\documentclass[dvipsnames]{article}

\usepackage[final]{openbmb_2025}

\usepackage[utf8]{inputenc}
\usepackage[T1]{fontenc}
\usepackage{CJKutf8}

\usepackage{amsmath, amssymb, amsfonts, mathtools, amsthm}
\usepackage{nicefrac}  

\usepackage{url}

\usepackage{booktabs}    
\usepackage{multirow}    
\usepackage{makecell}    
\usepackage{tabularx}    
\usepackage{adjustbox}   
\usepackage{colortbl}    

\usepackage{graphicx}
\usepackage{svg}
\usepackage{wrapfig}
\usepackage{subcaption}
\usepackage{xcolor}
\usepackage{microtype}   

\usepackage{titlesec}
\usepackage{enumitem}    
\usepackage{pifont}      

\usepackage{pythonhighlight}
\usepackage{fancyvrb}
\usepackage{listings}
\usepackage[most]{tcolorbox}
\tcbuselibrary{listingsutf8}

\usepackage{enumitem}
\setitemize{itemsep=10pt,topsep=0pt,parsep=0pt,partopsep=0pt}

\newcolumntype{Y}{>{\normalfont\ttfamily\raggedright\arraybackslash}X}

\newcommand{\myit}[1]{\textnormal{\textit{#1}}}

\usepackage{color, soul}
\usepackage[most,breakable]{tcolorbox}

\usepackage[labelfont=bf]{caption}
\usepackage{enumitem}
\usepackage{sectsty}

\usepackage{fancyhdr}
\renewcommand{\headrulewidth}{1pt}
\makeatletter
\def\headrule{{\if@fancyplain\let\headrulewidth\plainheadrulewidth\fi
\hrule\@height\headrulewidth\@width\textwidth \vskip-\headrulewidth}}
\makeatother

\definecolor{BMBDarkBlue}{HTML}{315EFE}
\definecolor{BMBLightBlue}{HTML}{00D3ED}
\usepackage[colorlinks, linkcolor=BMBDarkBlue, anchorcolor=BMBDarkBlue, citecolor=BMBDarkBlue, urlcolor=BMBDarkBlue]{hyperref}

\sectionfont{\color{BMBDarkBlue}\fontfamily{zi4}\selectfont}
\subsectionfont{\color{BMBDarkBlue}\fontfamily{zi4}\selectfont}
\subsubsectionfont{\color{BMBDarkBlue}\fontfamily{zi4}\selectfont}

\newtcolorbox{mytheorem}{
  colback=gray!5,       
  colframe=gray!80,     
  boxrule=0.5pt,        
  arc=4pt,              
  left=4pt,             
  right=4pt,            
  top=4pt,              
  bottom=4pt,           
}

\newcommand{\fancyheadname}{\textit{\textbf{AgentCPM-GUI}}}

\title{AgentCPM-GUI: Building Mobile-Use Agents with Reinforcement Fine-Tuning}

\author{
Zhong Zhang\textsuperscript{\rm 1}\thanks{\hspace{1mm}Equal contribution.}\hspace{1.5mm},
Yaxi Lu\textsuperscript{\rm 1}\footnotemark[1]\hspace{1.5mm},
Yikun Fu\textsuperscript{\rm 1}\thanks{\hspace{1mm}Work conducted during internships at Tsinghua University.}\hspace{1.5mm},
Yupeng Huo\textsuperscript{\rm 2},
Shenzhi Yang\textsuperscript{\rm 2},
Yesai Wu\textsuperscript{\rm 1},
Han Si\textsuperscript{\rm 1}\footnotemark[2]\hspace{1.5mm} \\
\textbf{Xin Cong\textsuperscript{\rm 1},
Haotian Chen\textsuperscript{\rm 1},
Yankai Lin\textsuperscript{\rm 2}\thanks{\hspace{1mm}Corresponding authors.}\hspace{1.5mm}, 
Jie Xie\textsuperscript{\rm 1},
Wei Zhou\textsuperscript{\rm 1},
Wang Xu\textsuperscript{\rm 1},
Yuanheng Zhang\textsuperscript{\rm 1}\footnotemark[2]} \\
\textbf{Zhou Su\textsuperscript{\rm 3},
Zhongwu Zhai\textsuperscript{\rm 3},
Xiaoming Liu\textsuperscript{\rm 3},
Yudong Mei\textsuperscript{\rm 3},
Jianming Xu\textsuperscript{\rm 3},
Hongyan Tian\textsuperscript{\rm 3}} \\
\textbf{Chongyi Wang\textsuperscript{\rm 3},
Chi Chen\textsuperscript{\rm 1},
Yuan Yao\textsuperscript{\rm 1},
Zhiyuan Liu\textsuperscript{\rm 1}\footnotemark[3]\hspace{1.5mm},
Maosong Sun\textsuperscript{\rm 1}}\footnotemark[3] \\
\textsuperscript{\rm 1}Tsinghua University ~~ 
\textsuperscript{\rm 2}Renmin University of China ~~ 
\textsuperscript{\rm 3}ModelBest Inc. \\
\texttt{zhongzhang@tsinghua.edu.cn}\quad
\texttt{lyx23@mails.tsinghua.edu.cn}
}

\def\huggingface{\raisebox{-2.4pt}{\includegraphics[height=1.05em]{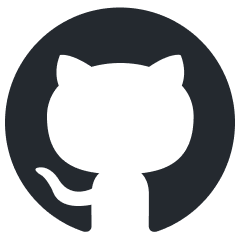}}}
\newcommand{\hflink}{https://github.com/OpenBMB/AgentCPM-GUI}

\begin{document}

\maketitle
\thispagestyle{fancy} 

\vspace{1em}
\begin{abstract}
The recent progress of large language model agents has opened new possibilities for automating tasks through graphical user interfaces (GUIs), especially in mobile environments where intelligent interaction can greatly enhance usability. However, practical deployment of such agents remains constrained by several key challenges. Existing training data is often noisy and lack semantic diversity, which hinders the learning of precise grounding and planning. Models trained purely by imitation tend to overfit to seen interface patterns and fail to generalize in unfamiliar scenarios. Moreover, most prior work focuses on English interfaces while overlooks the growing diversity of non-English applications such as those in the Chinese mobile ecosystem. In this work, we present AgentCPM-GUI, an 8B-parameter GUI agent built for robust and efficient on-device GUI interaction. Our training pipeline includes grounding-aware pre-training to enhance perception, supervised fine-tuning on high-quality Chinese and English trajectories to imitate human-like actions, and reinforcement fine-tuning with GRPO to improve reasoning capability. We also introduce a compact action space that reduces output length and supports low-latency execution on mobile devices. AgentCPM-GUI achieves state-of-the-art performance on five public benchmarks and a new Chinese GUI benchmark called CAGUI, reaching $96.9\%$ Type-Match and $91.3\%$ Exact-Match. To facilitate reproducibility and further research, we publicly release all code, model checkpoint, and evaluation data.
\end{abstract}

\section{Introduction}
The rapid advancements in Large Language Models (LLMs) and Multimodal Large Models (MLLMs) have catalyzed a new era of autonomous AI agents~\citep{LLMSurvey,survey_wang}. These agents are increasingly capable of understanding complex instructions~\citep{DBLP:conf/nips/Ouyang0JAWMZASR22,DBLP:conf/acl/QianHZDQCZZL0024}, performing multi-step planning~\citep{DBLP:journals/corr/abs-2402-02716}, and interacting with external tools or environments~\citep{DBLP:conf/iclr/QinLYZYLLCTQZHT24,DBLP:journals/csur/QinHLCDCZZHXHFSWQTZLSXZ25}. A critical frontier for deploying these intelligent agents in practical, human-centric applications is enabling them to proficiently operate Graphical User Interfaces (GUIs)~\citep{DBLP:journals/corr/abs-2411-04890,DBLP:journals/corr/abs-2412-13501,DBLP:journals/corr/abs-2411-18279}, particularly within the ubiquitous Android ecosystem, where they serve as the primary interaction layer for a vast array of daily digital tasks. Empowering LLM agents to seamlessly navigate and manipulate these mobile GUIs is essential for transforming them into truly versatile digital assistants capable of automating a wide spectrum of tasks on smartphones, thereby enhancing user productivity and accessibility. 

Early GUI agents emerged when Vision-Language Models (VLMs) had limited ability in reliably control GUI widgets. To compensate, researchers augmented model inputs with structured metadata, such as Android view hierarchies and system APIs, and even off-loaded perception and planning to more capable external VLMs (e.g., GPT-4o~\citep{openai2024gpt4o}), thereby improving widget grounding and action execution~\citep{DBLP:conf/chi/ZhangYLLHCHF025,DBLP:journals/corr/abs-2404-01549,DBLP:journals/corr/abs-2404-01744,DBLP:conf/icml/ZhengGK0024,DBLP:conf/nips/KimBM23,DBLP:journals/corr/abs-2401-16158}. Although effective, these hybrid pipelines propagated errors from cross-modal mismatches, incurred round-trip latency, and depended on metadata that many apps do not expose, creating significant challenges for generality and scalability. Recent GUI agents have advanced to resolving interface elements directly from raw pixels, enabling a single end-to-end model to match or even surpass earlier hybrid approaches~\citep{cogagent,DBLP:conf/acl/ChengSCX0Z024,uitars,aguvis,osatlas,showui,DBLP:conf/acl/0001Z24}. This shift positions purely visual, end-to-end modeling as the most scalable paradigm.

Despite significant progress, current visual GUI agents still face several challenges:
\textbf{(1) Data quality and scale}. High-quality, fine-grained interaction trajectories that capture realistic user behavior in diverse mobile apps are notoriously difficult to collect at scale. Most publicly available datasets either rely on synthetic generation or emulator-based recordings, both of which can introduce noise and lack semantic diversity. Such imperfect supervision limits the agent's ability to learn precise widget grounding, compositional reasoning, and long-horizon action planning.
\textbf{(2) Reasoning generalization}. GUI agents that are trained solely via imitation learning tend to overfit to interface patterns, resulting in brittle planning and poor generalization when task instructions deviate from seen templates or when UI layouts exhibit minor variations.
\textbf{(3) Language and regional coverage}. Current research concentrates almost exclusively on English GUIs, paying limited attention to the rapidly growing and diverse Chinese mobile ecosystem, whose interface design conventions and linguistic cues differ substantially. These differences limit the generalizability of current agents in multilingual and culturally diverse settings.

To address these challenges, we propose AgentCPM-GUI, a VLM-based agent for mobile GUI understanding and interaction. The key features of this work are as follows.

\begin{itemize}

\item \textbf{High-quality training data.}
We curate a large-scale corpus of $55$K trajectories with $470$K steps, encompassing a wide variety of Chinese Android apps via targeted collection and meticulous annotation. To enhance generalization and mitigate overfitting, we further incorporate and rigorously de-duplicate multiple public English Android datasets. The resulting unified dataset supports effective training, enabling robust cross-lingual and cross-app behavior modeling.

\item \textbf{Progressive training for perception, imitation, and reasoning.}
We adopt a three-stage progressive training pipeline to equip the agent with strong GUI understanding and reasoning capabilities, consisting of grounding-aware pre-training to enhance visual perception; supervised fine-tuning (SFT) to establish a reliable behavioral prior; and reinforcement fine-tuning (RFT)~\citep{rft-openai,DBLP:journals/corr/abs-2402-03300,DBLP:conf/acl/TrungZJSJL24} to further strengthen reasoning ability, enabling robust performance on long-horizon and compositional tasks. In addition, we optimize the training framework with asynchronous rollout and load balancing to support scalable reinforcement learning.

\item \textbf{Edge device oriented design.}  
To reduce decoding overhead, we carefully select action tokens to avoid unnecessary token fragmentation and adopt a compact JSON-based action format, resulting in an average output length of just $9.7$ tokens per action. While prior works largely overlook redundancy in action space design, our concise representation significantly improves runtime efficiency, enabling smooth and responsive on-device execution. 

\item \textbf{Comprehensive benchmarking.}  
We evaluate AgentCPM-GUI on the widely used English GUI agent benchmarks: AndroidControl~\citep{DBLP:journals/corr/abs-2406-03679}, GUI-Odyssey~\citep{DBLP:journals/corr/abs-2406-08451}, and AITZ~\citep{aitz}. In addition, we introduce \textbf{CAGUI}, the first large-scale \underline{C}hinese \underline{A}ndroid \underline{GUI} benchmark. CAGUI is a representative subset of our corpus designed for public evaluation. AgentCPM-GUI achieves new state-of-the-art performance across all datasets, demonstrating robust multilingual and cross-app generalization.
\end{itemize}

To support community research and ensure reproducibility, we open-source all model, training and evaluation code, along with the CAGUI benchmark. We believe this work establishes a solid foundation for advancing multilingual GUI agents in real-world applications.

\section{Related Work}

\subsection{Datasets and Benchmarks}

To facilitate progress in the field of GUI Agents, a number of datasets and benchmarks have been developed. Some of these datasets are specifically constructed for grounding tasks, which aim to associate natural language instructions with corresponding GUI widgets on the screen. Representative examples include Mind2Web~\citep{mind2web}, ScreenSpot/-Pro~\citep{DBLP:conf/acl/ChengSCX0Z024,DBLP:journals/corr/abs-2504-07981}, OS-ATLAS~\citep{osatlas}, GUICourse~\citep{DBLP:journals/corr/abs-2406-11317} and UGround~\citep{DBLP:conf/iclr/GouWZXCS0025}. Others are constructed for GUI agent tasks, where tasks are represented as sequences of action-observation pairs, such as AITW~\citep{androidinthewild}, AITZ~\citep{aitz}, AndroidControl~\citep{DBLP:journals/corr/abs-2406-03679}, GUI-Odyssey~\citep{DBLP:journals/corr/abs-2406-08451}, AMEX~\citep{DBLP:journals/corr/abs-2407-17490} and AndroidWorld~\citep{androidworld}. Despite their contributions, existing datasets predominantly focus on English GUIs, limiting the development of GUI agents in non-English environments. To bridge this gap, we introduce CAGUI, a new benchmark covering both grounding and agent tasks in realistic Chinese mobile apps, enabling more robust assessment in multilingual settings.

\subsection{VLM-based GUI Agents}

GUI agents have moved from API-based agent frameworks to vision-only, end-to-end agent models. Early work such as AppAgent~\citep{DBLP:conf/chi/ZhangYLLHCHF025,DBLP:journals/corr/abs-2408-11824} fused GPT-4(V)~\citep{2023GPT4VisionSC} with Android XML trees to build manuals for each screen. Mobile-Agent~\citep{DBLP:journals/corr/abs-2401-16158} and SeeAct~\citep{DBLP:conf/icml/ZhengGK0024} couple grounding modules with a reasoning LLM to decompose goals, ground elements and retry when actions fail. Newer agents place all perception inside a VLM that operates on raw screenshots~\citep{DBLP:conf/acl/0001Z24,DBLP:journals/corr/abs-2406-08451,showui,aria}. Recent efforts focus on scaling both models and data. CogAgent~\citep{cogagent}, Aguvis~\citep{aguvis} and UI-TARS~\citep{uitars} pre-train high-capacity (7B-70B) VLMs on millions of screenshots for GUI grounding, then fine-tune models on interaction trajectories to enhance planning and reasoning capabilities. Notably, OS-Atlas~\citep{osatlas} scales pre-training to $2.3$M cross-platform screens with $13$M labeled elements and releases the data and model. OS-Genesis~\citep{genesis} automates data collection by letting a seed agent explore apps, record its own steps, and filter them into high-quality trajectories used for later training.

\subsection{Reinforcement Learning for GUI Agents}

Most GUI agents rely on behavior cloning from fixed demonstrations, making their reasoning capability degenerate into pattern matching. Introducing reinforcement feedback lets agents learn from interaction and rewards, improving reasoning generalization. 

DigiRL~\citep{digirl} leverages a two-stage pipeline that first applies offline RL for initial policy learning, followed by online RL to allow the agent to improve through real-time exploration and feedback. 
DistRL~\citep{distrl} proposes a scalable asynchronous reinforcement learning framework for on-device agents, combining centralized policy training with decentralized data collection and a custom off-policy algorithm to improve training efficiency and task success in dynamic mobile environments.
Digi-Q~\citep{digiq} trains a Q-value function offline on frozen VLM representations and employs a best-of-N action sampling strategy to optimize policies without requiring additional environment interactions. 

Recently, RFT-tuned VLMs have shown promising performance on various vision tasks~\citep{DBLP:conf/nips/ZhaiBLPTZSXL0L24,DBLP:journals/corr/abs-2503-01785,DBLP:journals/corr/abs-2503-20752,DBLP:journals/corr/abs-2503-06749,DBLP:journals/corr/abs-2503-05132}, and extending this strategy to GUI agents has yielded notable improvements. UI-R1 and GUI-R1~\citep{uir1, guir1} adopt simple rule-based reward functions to assess the correctness of actions, using these signals to fine-tune the agent's capabilities through reinforcement. InfiGUI-R1~\citep{DBLP:journals/corr/abs-2504-14239} proposes a reasoning-centric progressive training paradigm, transforming GUI agents from reactive executors into deliberative reasoners. AppVLM~\citep{appvlm} adopts an RFT framework that iteratively collects successful trajectories through online interaction and refines the policy via supervised fine-tuning, enabling efficient policy improvement without relying on reinforcement learning algorithms.

\section{Method}

\subsection{Architecture Overview}

\begin{figure}[t]
    \centering
    \includegraphics[width=0.84\linewidth]{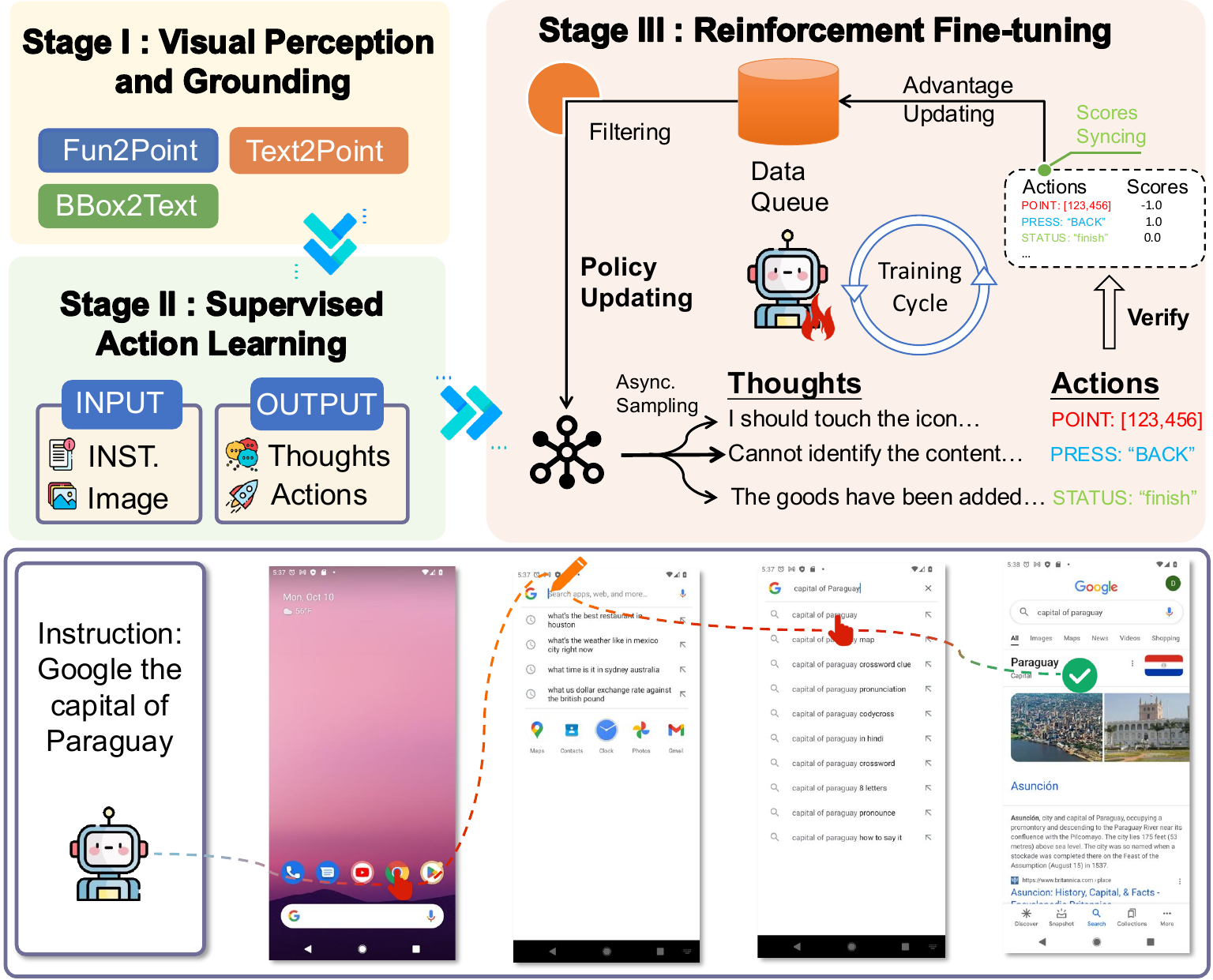}
    \caption{Overview of our training framework.}
    \label{fig:arch-pipeline}
\end{figure}

To train a VLM capable of performing GUI-based interactions, we adopt a three-stage training framework that incrementally builds the model's capabilities from perception to action. Each stage targets a distinct subskill essential for robust and generalizable GUI operation. Our GUI agent is built upon MiniCPM-V~\citep{DBLP:journals/corr/abs-2408-01800}, a lightweight vision-language model with 8B parameters that supports efficient visual grounding and instruction following, making it particularly well-suited for mobile-centric applications.

\paragraph{Stage I: Visual Perception and Grounding.} In the first stage, we focus on enhancing the model's perceptual and grounding abilities. We curate a dataset of vision-language alignment tasks, including OCR and widget-localization, to help the model learn fine-grained spatial and semantic correspondences between GUI widgets and their descriptions. This establishes a strong foundation for the subsequent training stages.

\paragraph{Stage II: Supervised Imitation Learning.} In the second stage, we collect GUI task execution trajectories paired with natural language instructions. Based on the model trained in Stage I, we conduct supervised fine-tuning to teach the model to generate valid and context-aware actions. This stage enables the model to imitate human-like action sequences when given a query, bridging the gap between perception and action.

\paragraph{Stage III: Reinforcement Fine-tuning.}
In the final stage, we apply RFT to further improve the model's reasoning and decision-making capabilities in complex GUI environments. Using the collected trajectories as initial demonstrations, we train the model with Group Relative Policy Optimization (GRPO)~\citep{DBLP:journals/corr/abs-2402-03300}. This process optimizes the model's reasoning and action capabilities by rewarding correct and goal-directed action sequences, thereby pushing the model beyond simple imitation into more robust autonomous planning and adaptive behavior.

This staged curriculum progressively transforms a general-purpose vision-language model into a GUI-capable agent, leveraging a structured combination of perception-level training, supervised action learning, and reinforcement-based policy optimization.

\subsection{Action Space Design}

A well-designed, unified, and language model-friendly action space is crucial for enabling models to understand and generalize behaviors effectively, as highlighted in numerous prior works. Inspired by these studies, we propose an action space that both reduces generation length and supports compositional actions. Our action space consists of six primary atomic action types listed as follows, and example actions are listed in Table~\ref{action_example}.

\begin{itemize}[left=0pt,]
\item \texttt{POINT}: Allows the model to locate a coordinate for performing operations. It takes a tuple \texttt{(x,y)} of integers in the range \texttt{[0,1000]}, normalized with \texttt{[0,0]} at the top-left corner of the current window and \texttt{[1000,1000]} at the bottom-right. By default, this action performs a tap at the specified coordinate. It can also be combined with \texttt{to} or \texttt{duration} to express a swipe gesture or a long-press action, respectively.

\item \texttt{to}: Enables scrolling in the window. This action can either specify a direction from \texttt{\{\text{"up"}, \text{"down"}, \text{"left"}, \text{"right"}\}} or define a swipe gesture when combined with a \texttt{POINT} target coordinate.

\item \texttt{TYPE}: Allows the model to input text into the current window. This action takes a string as its argument.

\item \texttt{PRESS}: Triggers special device keys, including \texttt{"HOME"}, \texttt{"BACK"}, \texttt{"ENTER"}. These keys represent common actions across devices and provide a concise way to express frequent operations.

\item \texttt{STATUS}: Enables the model to update the current task status, including \texttt{"continue"}, \texttt{"finish"}, \texttt{"satisfied"}, \texttt{"impossible"}, \texttt{"interrupt"}, \texttt{"need\_feedback"}. This action is used to communicate the state of execution. The default status is \texttt{"continue"} and can be omitted.

\item \texttt{duration}: Specifies the length of time (in milliseconds) that an action should persist. This parameter can be used independently to express an idle wait or combined with other actions (e.g., \texttt{POINT}) to indicate a long press or swipe duration.

\end{itemize}
To enhance efficiency and reduce token overhead during generation, we adopt a compact JSON representation that eliminates unnecessary whitespace between control characters. This contributes to a low average token cost of $9.7$ per action, helping to reduce latency and improve responsiveness on edge devices.

\begin{table}[ht]
\centering
\caption{Example actions of AgentCPM-GUI.}
\label{action_example}
\begin{adjustbox}{max width=\textwidth}
\begin{tabularx}{\textwidth}{lX}
\toprule
\textbf{Example Actions} & \textbf{Purpose} \\
\midrule
\verb|{"POINT":[480,320]}| & Single tap at the normalized screen coordinate. \\
\verb|{"POINT":[480,320],"duration":1000}| & Touch-and-hold (long press) at the coordinate. \\
\verb|{"POINT":[500,200],"to":"down"}| & Swipe to a direction or another coordinate. \\
\verb|{"PRESS":"HOME"}| & Trigger a hardware or navigation key. \\
\verb|{"TYPE":"Hello, world!"}| & Insert the given text at the current input focus. \\
\verb|{"duration":500}| & Idle for the specified time in milliseconds. \\
\verb|{"STATUS":"finish"}| & Task status, can stand alone or with an action. \\
\bottomrule
\end{tabularx}
\end{adjustbox}
\end{table}

\subsection{Stage I: Visual Perception and Grounding}

For grounding pre-training, we collect Android GUI data by sampling examples from several open-source corpora (AITZ~\citep{aitz}, GUICourse~\citep{DBLP:journals/corr/abs-2406-11317}, OS-Atlas~\citep{osatlas}, UGround~\citep{DBLP:conf/iclr/GouWZXCS0025}, ScreenSpot~\citep{DBLP:conf/acl/ChengSCX0Z024}) and additional screenshots from our collected Chinese app data. Each image is formulated as either an OCR task that asks the model to write the text in a marked region, or a widget-localization task that asks it to output the bounding box coordinate of a referenced UI element. Grounding batches mix in $50\%$ general multimodal SFT data (e.g., Chat, VQA, Multimodal Reasoning)~\citep{DBLP:journals/corr/abs-2408-01800}, which regularizes the vision module while letting it absorb GUI-specific cues. In total, the grounding pre-training dataset comprises $12$M samples. 

This pre-training stage plays a crucial role in establishing the model's low-level perceptual and grounding abilities. We observe that, after this stage, the model demonstrates strong proficiency in identifying and locating GUI widgets, especially in accurately predicting coordinates based on visual cues. However, the model at this point still struggles to generate well-formed function calls or to reason over action types, indicating limited understanding of higher-level task semantics and planning. These capabilities are further enhanced in the subsequent SFT and RFT stages.

\subsection{Stage II: Supervised Imitation Learning}

Due to the scarcity of high-quality open-source datasets for Chinese Android apps, we constructed a large-scale, high-fidelity dataset of GUI interaction trajectories to support supervised imitation learning. The corpus covers over $30$ mainstream Chinese apps, spanning eight functional domains: life services, e-commerce, navigation, social, video, music/audio, reading/learning, and productivity. This ensures that the agent is exposed to a wide spectrum of UI layouts, widget types, and task intents. In total, we obtained $55$K complete task trajectories comprising $470$K atomic steps, approximately $8.5$ steps per trajectory. The data curation process is as follows:

\begin{itemize}[left=0pt,]
\item \textbf{Query Generation}. We first designed parameterized instruction templates for each app to reflect its core user intents. Slot values such as store names and quantities were then filled using GPT-4o. Human annotators reviewed the resulting queries, removing errors and duplicates. Finally, another GPT-4o pass paraphrased the verified queries into diverse surface forms, broadening lexical coverage and mitigating over-fitting.

\item \textbf{Trajectory Collection}. We collected trajectories on physical Android phones using a custom data logger. Running on real devices bypassed the feature limitations of emulators and preserved sensor signals like GPS and accelerometer data. Annotators received a queued query, completed the task, and confirmed each deliberate action. The logger saved only these validated taps, long-presses, swipes, text inputs, and their associated UI metadata while removing the spurious events that screen-capture pipelines often introduce.

\item \textbf{Quality Assurance}. Publicly available Android GUI corpora like AITW~\citep{androidinthewild} and AndroidControl~\citep{DBLP:journals/corr/abs-2406-03679} contain non-negligible mislabeled actions and other annotation errors introduced by unrestricted screen-recording pipelines. To ensure data quality, we implemented two measures. First, our data logger records actions only if they are explicitly confirmed by annotators, preventing accidental gestures from being logged. Second, we apply post-hoc filtering to remove trajectories that miss essential steps, fail to complete the intended task, or duplicate existing examples.
\end{itemize}

To warm up the model for reasoning, we introduced preliminary thought generation at the SFT stage. We annotated $24$K interaction trajectories from Chinese apps using GPT-4o to supply intermediate reasoning supervision. The dataset was further augmented with English thought-annotated data from AITZ~\citep{aitz} and AndroidControl~\citep{DBLP:journals/corr/abs-2406-03679}. This warm-up is essential because without it, the model failed to generate reasoning traces during the RFT stage. In addition, to enable controlled thought generation at inference time and the RFT stage, samples with thought traces adopted a schema where the \texttt{thought} field was marked as "\texttt{required}", whereas those without were marked as "\texttt{optional}".

In order to enhance cross-lingual generalization and reduce over-fitting, we augmented our Chinese corpus with publicly available English-language datasets: AITW~\citep{DBLP:journals/corr/abs-2406-03679}, AITZ~\citep{aitz}, AMEX~\citep{DBLP:journals/corr/abs-2407-17490}, AndroidControl~\citep{DBLP:journals/corr/abs-2406-03679}, and GUI-Odyssey~\citep{DBLP:journals/corr/abs-2406-08451}. Since AITW is internally redundant, we performed intra-query de-duplication. For each trajectory, we extracted ResNet-50 features from its screenshots and averaged them to produce a trajectory embedding. We then grouped trajectories by shared query and, within each group, removed those whose cosine similarity to any previously retained sample exceeded a fixed threshold. This retained approximately $40\%$ of the original data.

Empirically, training solely on GUI-interaction data led to a pronounced mode collapse during the subsequent RFT stage, manifesting as impoverished and repetitive reasoning thoughts. To mitigate this, we mixed $50\%$ general multimodal SFT data into training batches, which helped stabilize policy optimization. The SFT data comprises a mix of single-turn (system-user-assistant) and multi-turn dialogues. For multi-turn examples, we retained only the last three turns of user-assistant interaction to provide sufficient conversational context while keeping input sequences within tractable length limits. In total, $6.9$M instances were used for the SFT stage.

\subsection{Stage III: Reinforcement Fine-tuning}

We introduce an RFT stage to improve the agent's reasoning ability. To make RFT practical at scale, we further develop a training framework which supports asynchronous rollout and two levels of load balancing to improve efficiency and scalability across distributed environments.

\subsubsection{Algorithmic Design}

\paragraph{GRPO Algorithm.} We conduct RFT based on the GRPO~\citep{DBLP:journals/corr/abs-2402-03300} algorithm. GRPO replaces the value critic of PPO~\citep{DBLP:journals/corr/SchulmanWDRK17}
with a
group-wise comparison of candidate completions.  
Given a query \(q\), the current policy
\(\pi_{\theta_{\mathrm{old}}}\) samples \(N\) responses
\(\{o_1,\dots,o_N\}\).  Each response is assigned a scalar
task reward \(\{r_1,\dots,r_N\}\).
Rewards are normalised within the group to produce variance-reduced advantages:
\begin{equation}
\hat{A}_i=\frac{r_i-\operatorname{mean}(\mathbf r)}
          {\operatorname{std}(\mathbf r)},\quad
\mathbf r=\{r_1,\dots,r_N\}.
\label{eq:grpo-adv}
\end{equation}

The policy is then updated using a clipped objective with KL divergence penalty:
\begin{equation}
\begin{split}
    \mathcal{J}_{\text{GRPO}}(\theta) = \mathbb{E}_{q \sim P(Q), \{o_i\}_{i=1}^G \sim \pi_{\theta_{old}}(O|q)}
    \Bigg[ \frac{1}{G} \sum_{i=1}^G \frac{1}{|o_i|} \sum_{t=1}^{|o_i|} \Bigg\{ \min \Bigg[
    \frac{\pi_\theta(o_{i,t} | q, o_{i,<t})}{\pi_{\theta_{old}}(o_{i,t} | q, o_{i,<t})} \hat{A}_{i,t},
    \quad \\ \text{clip} \left( \frac{\pi_\theta(o_{i,t} | q, o_{i,<t})}{\pi_{\theta_{old}}(o_{i,t} | q, o_{i,<t})}, 1 - \epsilon, 1 + \epsilon \right)\hat{A}_{i,t}
    \Bigg] - \beta \mathbb{D}_{KL}\left[\pi_{\theta} || \pi_{ref}\right] \Bigg\} \Bigg],
\end{split}
\label{eq:GRPO-obj}
\end{equation}
where $\pi_{\theta}$ and $\pi_{\theta_{old}}$ are the current and old policy, and $\epsilon$ and $\beta$ are hyperparameters.

\paragraph{Reward Design and Validation.} During RFT, we apply a two-stage validation scheme to evaluate model outputs: (1) format checking and (2) semantic correctness. The reward is mapped to the range $[- 1,1]$. If an output fails the format check (e.g., malformed structure or missing fields), a reward of $- 1$ is assigned. If the format is correct but the answer is semantically incorrect, the reward is $0$. If both format and answer are correct, the reward is $1$. For action spaces involving continuous goals, such as predicting a \texttt{POINT} target, we further define correctness by spatial accuracy: if the predicted point falls within the ground-truth bounding box, a reward of $1$ is assigned; otherwise, $0$. This fine-grained reward design encourages both syntactic correctness and task-specific accuracy.

\subsubsection{System Optimization}

Our training system adopts an asynchronous architecture that decouples rollout execution from policy updates. Once a task ID is dispatched from the global task queue, it is sampled $n$ times according to the GRPO algorithm to generate multiple candidate responses per policy. After inference and reward computation for each sample are complete, the main process computes the advantage for the samples using GRPO's variance-reduced estimator. These advantage values are then sent to the node-level main process for policy updating. The global main process collects all necessary statistics and, when synchronization conditions are met, coordinates a unified policy update across nodes. This design ensures tight integration of GRPO's optimization logic within our distributed, asynchronous training framework.

\paragraph{Asynchronous Rollout.} In our design, each GPU group performs inference independently and asynchronously. The inference results are first synchronized to the local node's main process. Then, each local main process communicates its inference status with a global main process, which tracks global rollout progress and coordinates training updates. During inference, each GPU group also asynchronously requests the next batch of data required for computing policy gradients. The global main process monitors the overall rollout status and, once a pre-defined synchronization condition is met, broadcasts a signal to all GPU groups to pause rollout and perform a synchronized model update. This asynchronous rollout scheme ensures that GPU groups operate efficiently without waiting for each other, thus fully utilizing computational resources.

\paragraph{Hierarchical Load Balancing.} The asynchronous design introduces challenges related to load imbalance, particularly at two levels: intra-node (between GPU groups) and inter-node (between different compute nodes). Intra-node imbalance is addressed by constructing a global task queue from which inference tasks are dynamically dispatched to GPU groups. This design make each GPU group consistently have access to available tasks, thereby minimizing idle time. However, nodes with differing hardware configurations or system loads can result in inter-node imbalance: some nodes may accumulate more rollout results than others. To address this, we implement a work stealing mechanism: underutilized nodes can request inference results from overburdened peers. This approach is particularly suited for large-scale, multi-modal inference outputs, which are often expensive to transmit and manage. Work stealing provides a flexible and scalable solution that avoids the drawbacks of forced synchronization across machines.

\section{Experiments}

\subsection{GUI Grounding Capability}

\begin{table}[b]
    \centering
    \caption{GUI grounding accuracy on the CAGUI benchmark over the Fun2Point, Text2Point, and Bbox2Text sub-tasks. \textbf{Bold} and \underline{underline} indicate the best and second-best results.}
    \label{tab:gui_grounding}
    \begin{adjustbox}{max width=\textwidth}
    \begin{tabularx}{\textwidth}{l
                                *{4}{>{\centering\arraybackslash}X}}
    \toprule
    \textbf{Models} & \textbf{Fun2Point} & \textbf{Text2Point} & \textbf{Bbox2Text} & \textbf{Average} \\
    \midrule
    \multicolumn{5}{c}{\textbf{\textit{Closed-source Models}}} \\
    \midrule
    GPT-4o~\citep{openai2024gpt4o} & 22.1 & 19.9 & 14.3 & 18.8 \\
    GPT-4o with grounding~\citep{DBLP:journals/corr/abs-2408-00203} & 44.3 & 44.0 & 14.3 & 34.2 \\

    \midrule
    \multicolumn{5}{c}{\textbf{\textit{Open-source Models}}} \\
    \midrule
    
    Qwen2.5-VL-7B~\citep{qwenvl} & 59.8 & 59.3 & \underline{50.0} & \underline{56.4} \\
    InternVL2.5-8B~\citep{dong2024internlm} & 17.2 & 24.2 & 45.9 & 29.1 \\
    InternVL2.5-26B~\citep{dong2024internlm} & 14.8 & 16.6 & 36.3 & 22.6 \\
    OS-Genesis-7B~\citep{genesis} & 8.3 & 5.8 & 4.0 & 6.0 \\
    UI-TARS-7B~\citep{uitars} & 56.8 & \underline{66.7} & 1.4 & 41.6 \\
    OS-Altas-7B~\citep{osatlas} & 53.6 & 60.7 & 0.4 & 38.2 \\
    Aguvis-7B~\citep{aguvis} & \underline{60.8} & \textbf{76.5} & 0.2 & 45.8 \\
    \midrule
    \rowcolor{blue!10}
    AgentCPM-GUI & \textbf{79.1} & \textbf{76.5} & \textbf{58.2} & \textbf{71.3} \\
    \bottomrule
    \end{tabularx}
    \end{adjustbox}
\end{table}

We evaluate GUI grounding on CAGUI through three tasks designed to assess different aspects of visual-language alignment and understanding: \textbf{1) Fun2Point}. Given a description of a component's function in the GUI (e.g., "this button opens the website"), the model must locate the correct coordinates of the mentioned component; \textbf{2) Text2Point}. The model is required to locate a given textual string appearing within the GUI; \textbf{3) Bbox2Text}. The model receives a bounding box location on the GUI and must accurately output the corresponding textual content. Representative examples of these tasks are included in Appendix~\ref{appendix:CAGUI_Grounding_Task_Examples}.

All three grounding tasks are evaluated on the CAGUI benchmark, which was specifically curated for assessing GUI grounding capability in Chinese Android apps. The raw dataset consists of screenshots paired with corresponding XML metadata collected from real-world apps. Each XML file provides fine-grained annotations for GUI widgets, including bounding box coordinates, textual content, and component types.
For the Text2Point and Bbox2Text tasks, annotations were directly extracted from the XML metadata by aligning textual content with their corresponding bounding boxes. For Fun2Point, additional function-level labels were constructed to reflect the semantic roles of GUI widgets. To generate these labels, we first overlaid bounding boxes onto the screenshots to explicitly highlight the spatial boundaries of each widget. Then, we prompted a strong VLM Qwen2.5-VL-72B to produce concise functional descriptions, yielding high-quality semantic labels for each widget.

Evaluation procedures were tailored to the input-output formats of each model. InternVL models output bounding boxes, which are evaluated against the ground-truth using the Intersection-over-Union (IoU) metric, with a threshold of $0.5$ indicating a successful match. GPT-4o is augmented with OmniParser~\citep{DBLP:journals/corr/abs-2408-00203}, which extracts layout structures and text/icon segments before the model predicts a target box index. Models including ours generate point coordinates and are assessed by comparing them with ground-truth locations under a predefined spatial tolerance.

The results are summarized in Table~\ref{tab:gui_grounding}. AgentCPM-GUI significantly outperforms all baselines across all three tasks. In particular, it achieves a large performance margin in the Bbox2Text task, where most baseline models struggle-largely due to the need for precise alignment between visual regions and text content. Despite the task's difficulty, AgentCPM-GUI attains a $58.2\%$ accuracy, while nearly all competing models score below $5\%$. This highlights our model's superior grounding ability, especially in mobile interface contexts where visual complexity, small text, and overlapping elements pose unique challenges.

\subsection{Action Prediction Capability}

\begin{table}[htpb]
    \centering
    \caption{Step-level action prediction performance on five GUI Agent benchmarks, in terms of Type Match (TM) and Exact Match (EM). \textbf{Bold} and \underline{underline} indicate the best and second-best results. *OS-Atlas uses different train/test splits on GUI-Odyssey benchmark and is not directly comparable.}
    \label{tab:gui_evaluation}
    \begin{adjustbox}{max width=\textwidth}
    \begin{tabularx}{\textwidth}{l
                                *{10}{>{\centering\arraybackslash}X}}
    \toprule
    \multirow{2}{*}{\textbf{Models}} 
    & \multicolumn{2}{c}{\textbf{AC-Low}} 
    & \multicolumn{2}{c}{\textbf{AC-High}} 
    & \multicolumn{2}{c}{\textbf{Odyssey}} 
    & \multicolumn{2}{c}{\textbf{AITZ}} 
    & \multicolumn{2}{c}{\textbf{CAGUI}} \\
    \cmidrule(lr){2-3} 
    \cmidrule(lr){4-5} 
    \cmidrule(lr){6-7} 
    \cmidrule(lr){8-9} 
    \cmidrule(lr){10-11}
    & TM & EM & TM & EM & TM & EM & TM & EM & TM & EM \\
    \midrule
    \multicolumn{11}{c}{\textbf{\textit{Closed-source Models}}} \\
    \midrule

    GPT-4o~\citep{openai2024gpt4o}                   & -     & 19.5 & -     & 20.8 & -     & 20.4 & 70.0 & 35.3 & 3.67  & 3.67  \\
    Gemini 2.0~\citep{gemini20}               & -     & 28.5 & -     & 60.2 & -     & 3.27  & -     & -     & -     & -     \\
    Claude~\citep{claude35}                   & -     & 19.4 & -     & 12.5 & 60.9 & -     & -     & -     & -     & -     \\

    \midrule
    \multicolumn{11}{c}{\textbf{\textit{Open-source Models}}} \\
    \midrule
    
    Qwen2.5-VL-7B~\citep{qwenvl}            & 94.1 & 85.0 & 75.1 & 62.9 & 59.5 & 46.3 & 78.4 & 54.6 & 74.2 & 55.2 \\
    UI-TARS-7B~\citep{uitars}               & \textbf{95.2} & \textbf{91.8} & \textbf{81.6} & \textbf{74.4} & 86.1 & 67.9 & \underline{80.4} & \underline{65.8} & \underline{88.6} & \underline{70.3} \\
    OS-Genesis-7B~\citep{genesis}            & 90.7 & 74.2 & 65.9 & 44.4 & 11.7 & 3.63  & 20.0 & 8.45  & 38.1 & 14.5 \\
    OS-Atlas-7B~\citep{osatlas}             & 73.0 & 67.3 & 70.4 & 56.5 & 91.8* & 76.8* & 74.1 & 58.5 & 81.5 & 55.9 \\
    Aguvis-7B~\citep{aguvis}                & 93.9 & 89.4 & 65.6 & 54.2 & 26.7 & 13.5 & 35.7 & 19.0 & 67.4 & 38.2 \\
    OdysseyAgent~\citep{DBLP:journals/corr/abs-2406-08451}          & 65.1 & 39.2 & 58.8 & 32.7 & \underline{90.8} & \underline{73.7} & 59.2 & 31.6 & 67.6 & 25.4 \\

    \midrule
    \rowcolor{blue!10}
    AgentCPM-GUI & \underline{94.4} & \underline{90.2} & \underline{77.7} & \underline{69.2} & \textbf{90.9} & \textbf{75.0} & \textbf{85.7} & \textbf{76.4} & \textbf{96.9} & \textbf{91.3} \\

    \bottomrule
    \end{tabularx}
    \end{adjustbox}
\end{table}

We conduct a comprehensive evaluation of AgentCPM-GUI on representative benchmarks: AndroidControl~\citep{DBLP:journals/corr/abs-2406-03679}, GUI-Odyssey~\citep{DBLP:journals/corr/abs-2406-08451}, AITZ~\citep{aitz}, and CAGUI, covering diverse GUI interaction patterns across both English and Chinese environments. Each benchmark adopts two standard evaluation metrics: Type Match (TM), which checks if the predicted action type matches the ground truth, and Exact Match (EM), which additionally requires all parameters to be correctly predicted. As shown in Table~\ref{tab:gui_evaluation}, AgentCPM-GUI achieves state-of-the-art performance across all benchmarks. Notably, it demonstrates strong generalization in complex multi-step scenarios, such as those in GUI-Odyssey and AITZ, significantly outperforming existing models. On the CAGUI benchmark, our model achieves $96.9\%$ TM and $91.3\%$ EM, substantially ahead of other models, highlighting its effectiveness in Chinese-language GUI settings.

All baseline results are from our own re-implementations to ensure fair and reproducible comparisons. We closely followed each model's official instructions and prompts where available, and applied consistent input and evaluation protocols throughout. Notably, OS-Atlas uses a different train/test split on GUI-Odyssey benchmark, so its results are not directly comparable. Our evaluation code and benchmarks are publicly released to support reproducibility and future research.

\begin{table}[ht]
\centering
\caption{Ablation study comparing AgentCPM-GUI before and after RFT.}
\begin{adjustbox}{max width=\textwidth}
\begin{tabular}{lcccccccccc}
\toprule

\multirow{2}{*}{\textbf{Models}} 
& \multicolumn{2}{c}{\textbf{AC-Low}} 
& \multicolumn{2}{c}{\textbf{AC-High}} 
& \multicolumn{2}{c}{\textbf{Odyssey}} 
& \multicolumn{2}{c}{\textbf{AITZ}} 
& \multicolumn{2}{c}{\textbf{CAGUI}} \\
\cmidrule(lr){2-3} 
\cmidrule(lr){4-5} 
\cmidrule(lr){6-7} 
\cmidrule(lr){8-9} 
\cmidrule(lr){10-11}

& TM & EM & TM & EM & TM & EM & TM & EM & TM & EM \\
\midrule
AgentCPM-GUI-SFT & 87.6 & 83.1 & \textbf{78.6} & \textbf{69.5} & 86.1 & 66.7 & 79.0 & 61.1 & \textbf{96.9} & \textbf{91.5} \\
AgentCPM-GUI-RFT & \textbf{94.4} & \textbf{90.2} & 77.7 & 69.2 & \textbf{90.9} & \textbf{75.0} & \textbf{85.7} & \textbf{76.4} & \textbf{96.9} & 91.3 \\
\bottomrule
\end{tabular}
\end{adjustbox}
\label{tab:sft_rft_ablation}
\end{table}

\subsection{Effects of Reinforcement Fine-tuning}

To assess the contribution of RFT, we compare our model's performance before and after RFT across all benchmarks, as shown in Table~\ref{tab:sft_rft_ablation}. On challenging datasets such as AndroidControl-Low, GUI-Odyssey, and AITZ, RFT brought significant improvements, especially in exact match accuracy. This  demonstrates its effectiveness in enhancing the model's ability to handle long-horizon reasoning and complex decision-making. However, on datasets like AndroidControl-High and CAGUI, the SFT-only model already performed competitively or even slightly better. This is likely because these benchmarks have sufficiently large and diverse training sets, allowing the model to encounter similar patterns during supervised training. In such cases, imitation learning alone generalizes well, and additional reinforcement may yield limited benefit.

\begin{figure}[htpb]
    \centering
  \includegraphics[width=0.5\linewidth]{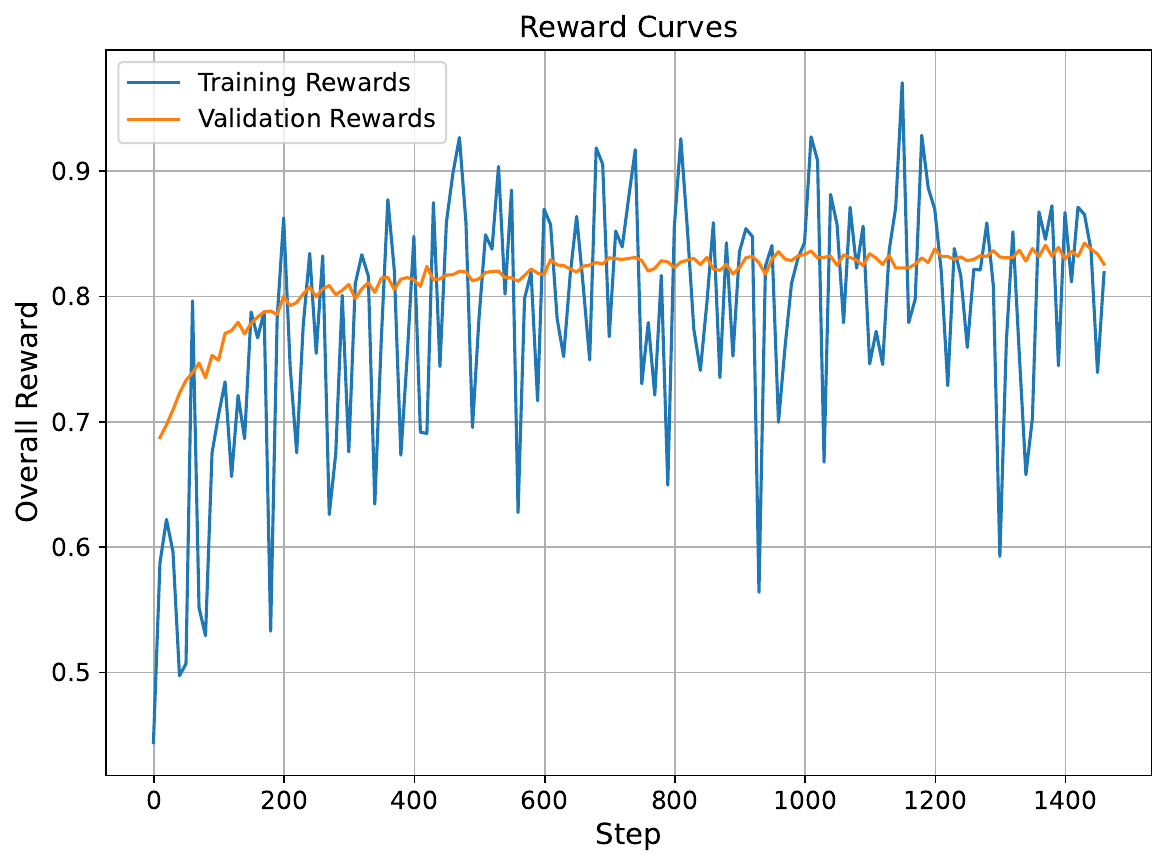}
    \caption{Reward curves on the training and validation sets of AgentCPM-GUI.}
    \label{fig:reward_curve}
\end{figure}

To monitor the optimization process, we held out a small subset of the training data as a validation set and tracked the reward curves on both the training and validation sets (Figure~\ref{fig:reward_curve}). Notably, the training reward curve shows substantial fluctuations since each point reflects a single mini-batch, while validation points are averaged over the entire held-out set. Despite the variance, the training reward trends upward, indicating effective learning. The validation reward rises steadily and plateaus around $0.82$, suggesting good generalization.

\section{Conclusion}

We present AgentCPM-GUI, a VLM-based agent designed for GUI interaction on mobile devices. Built upon MiniCPM-V, AgentCPM-GUI is trained through a three-stage pipeline to progressively acquire grounding, action, and reasoning capabilities. To support this process, we construct a high-quality Chinese Android interaction dataset and augment it with carefully selected open-source English data, enabling the agent to generalize effectively across both Chinese and English applications. In particular, our RFT stage equips the model with stronger reasoning and planning abilities, which are essential for handling complex, long-horizon GUI tasks. We also design a compact action space with an average output length of $9.7$ tokens, making the model well-suited for deployment on edge devices, which is a dimension often overlooked in previous work. Extensive experiments across public benchmarks and our newly introduced CAGUI benchmark demonstrate state-of-the-art performance, especially in Chinese-language environments. To foster further research and ensure reproducibility, we release all code, evaluation data, and model checkpoint.

\newpage

\bibliographystyle{citation}
\bibliography{citation}

\clearpage
\appendix

\section{Training Details}

We list the main hyperparameters for the SFT and RFT stages in Table~\ref{tab:sft_hyperparams} and Table~\ref{tab:training_parameters}, respectively.

\begin{table}[htpb]
\centering
\caption{Training parameters for Stage II: Supervised Fine-tuning.}
\label{tab:sft_hyperparams}
\begin{tabularx}{\textwidth}{c c X}
\toprule
\textbf{Parameter} & \textbf{Default Value} & \textbf{Description} \\
\midrule
\texttt{model\_max\_length} & 2304 & Maximum sequence length \\
\texttt{max\_line\_res} & 1120 & Maximum image resolution for the longest axis \\
\texttt{per\_device\_train\_batch\_size} & 1 & Training batch size per device \\
\texttt{gradient\_accumulation\_steps} & 1 & Gradient accumulation steps \\
\texttt{num\_train\_epochs} & 3 & Number of training epochs \\
\texttt{learning\_rate} & 1e-5 & Learning rate \\
\texttt{weight\_decay} & 0.1 & Weight decay coefficient \\
\texttt{adam\_beta1} & 0.9 & Adam optimizer beta1 parameter \\
\texttt{adam\_beta2} & 0.999 & Adam optimizer beta2 parameter \\
\texttt{max\_grad\_norm} & N/A & Gradient clipping disabled \\
\texttt{lr\_scheduler\_type} & cosine & Learning rate scheduler type \\
\texttt{warmup\_ratio} & 0.05 & Linear warmup ratio \\
\texttt{bf16} & True & Use bfloat16 precision \\
\texttt{gradient\_checkpointing} & False & Whether using gradient checkpointing \\
\texttt{deepspeed} & ZeRO-2 & Deepspeed optimization stage \\
\bottomrule
\end{tabularx}
\end{table}

\begin{table}[htpb]
\centering
\caption{Training parameters for Stage III: Reinforcement Fine-tuning.}
\begin{tabularx}{\textwidth}{c c X}
\toprule
\textbf{Parameter} & \textbf{Default Value} & \textbf{Description} \\
\midrule
\texttt{max\_prompt\_length} & 16384 & Maximum prompt length \\
\texttt{max\_completion\_length} & 512 & Maximum completion length \\
\texttt{max\_line\_res} & 1120 & Maximum image resolution for the longest axis \\
\texttt{num\_generations} & 8 & Number of generations \\
\texttt{per\_device\_train\_batch\_size} & 1 & Training batch size per device \\
\texttt{gradient\_accumulation\_steps} & 32 & Gradient accumulation steps \\
\texttt{learning\_rate} & 1e-6 & Learning rate \\
\texttt{num\_train\_epochs} & 3 & Number of training epochs \\
\texttt{weight\_decay} & 0.1 & Weight decay coefficient \\
\texttt{adam\_beta2} & 0.99 & Adam optimizer beta2 parameter \\
\texttt{max\_grad\_norm} & 1.0 & Maximum gradient norm for clipping \\
\texttt{lr\_scheduler\_type} & cosine & Learning rate scheduler type \\
\texttt{beta} & 0.04 & KL divergence coefficient \\
\texttt{bf16} & True & Use bfloat16 precision \\
\bottomrule
\end{tabularx}
\label{tab:training_parameters}
\end{table}

\clearpage

\section{Evaluation Details}

To ensure fair and consistent evaluation across all models, we adopt a unified evaluation framework. Since different models may define their own action formats and conventions, their outputs are first converted into a shared action representation defined by AgentCPM-GUI. This normalization allows us to compare models under the same evaluation criteria and metrics. In the following, we provide representative input prompts for each model, detail the evaluation settings and hyperparameters, and describe how action space conversion is performed when applicable.

\subsection{Qwen2.5-VL-7B}

\subsubsection{Data example}

\begin{tcolorbox}[breakable, colback=black!5!white,colframe=black!75!black,title=Qwen2.5-VL-7B Data Example]
\tcbsubtitle{System Message}
\begin{lstlisting}
You are a helpful assistant.

# Tools

You may call one or more functions to assist with the user query.

You are provided with function signatures within <tools></tools> XML tags:
<tools>
{"type": "function", "function": {"name_for_human": "mobile\_use", "name": "mobile\_use", "description": "Use a touchscreen to interact with a mobile device, and take screenshots.
* This is an interface to a mobile device with touchscreen. You can perform actions like clicking, typing, swiping, etc.
* Some applications may take time to start or process actions, so you may need to wait and take successive screenshots to see the results of your actions.
* The screen's resolution is 1092x2408.
* Make sure to click any buttons, links, icons, etc with the cursor tip in the center of the element. Don't click boxes on their edges unless asked.", "parameters": {"properties": {"action": {"description": "The action to perform. The available actions are:
* `key`: Perform a key event on the mobile device.
    - This supports adb's `keyevent` syntax.
    - Examples: \"volume\_up\", \"volume\_down\", \"power\", \"camera\", \"clear\".
* `click`: Click the point on the screen with coordinate (x, y).
* `long\_press`: Press the point on the screen with coordinate (x, y) for specified seconds.
* `swipe`: Swipe from the starting point with coordinate (x, y) to the end point with coordinates2 (x2, y2).
* `type`: Input the specified text into the activated input box.
* `system\_button`: Press the system button.
* `open`: Open an app on the device.
* `wait`: Wait specified seconds for the change to happen.
* `terminate`: Terminate the current task and report its completion status.", "enum": ["key", "click", "long\_press", "swipe", "type", "system\_button", "open", "wait", "terminate"], "type": "string"}, "coordinate": {"description": "(x, y): The x (pixels from the left edge) and y (pixels from the top edge) coordinates to move the mouse to. Required only by `action=click`, `action=long\_press`, and `action=swipe`.", "type": "array"}, "coordinate2": {"description": "(x, y): The x (pixels from the left edge) and y (pixels from the top edge) coordinates to move the mouse to. Required only by `action=swipe`.", "type": "array"}, "text": {"description": "Required only by `action=key`, `action=type`, and `action=open`.", "type": "string"}, "time": {"description": "The seconds to wait. Required only by `action=long\_press` and `action=wait`.", "type": "number"}, "button": {"description": "Back means returning to the previous interface, Home means returning to the desktop, Menu means opening the application background menu, and Enter means pressing the enter. Required only by `action=system\_button`", "enum": ["Back", "Home", "Menu", "Enter"], "type": "string"}, "status": {"description": "The status of the task. Required only by `action=terminate`.", "type": "string", "enum": ["success", "failure"]}}, "required": ["action"], "type": "object"}, "args\_format": "Format the arguments as a JSON object."}}
</tools>

For each function call, return a json object with function name and arguments within <tool\_call></tool\_call> XML tags:
<tool_call>
{"name": <function-name>, "arguments": <args-json-object>}
</tool_call>
\end{lstlisting}
\tcbsubtitle{User}
The user query:  [user\_request]

Current step query: {low\_lew\_instruction} (included only when low\_lew\_instruction is defined)

Task progress (You have done the following operation on the current device): [history\_actions]

[current\_screenshot]
\tcbsubtitle{Assistant}
[thought\_and\_action]
\label{eg_qwen}
\end{tcolorbox}

\subsubsection{Action Space Mapping}

Table~\ref{tab:qwen_mapping} shows the action space mapping from Qwen2.5-VL-7B to the standardized representation. Two key differences must be addressed during conversion. First, Qwen2.5-VL-7B expresses duration in seconds for actions such as \texttt{long\_press} and \texttt{wait}, whereas AgentCPM-GUI expects time in milliseconds. Second, Qwen2.5-VL-7B produces absolute screen coordinates (in pixels) for spatial actions like \texttt{click}, \texttt{long\_press}, and \texttt{swipe}, while AgentCPM-GUI uses normalized coordinates in the range $[0,1000]$ relative to screen size. 

\begin{table}[htbp]
\centering
\caption{Action space mapping from Qwen2.5-VL-7B to AgentCPM-GUI.}
\label{tab:qwen_mapping}
\begin{tabularx}{\textwidth}{l XY}
\toprule
\textbf{Qwen2.5-VL-7B} & \textbf{Input Parameters} & {\normalfont\textbf{AgentCPM-GUI}} \\
\midrule
\texttt{click} & coordinate = (x, y) &
\{"POINT":[int(x/width*1000),\newline int(y/height*1000)]\} \\

\texttt{long\_press} & coordinate = (x, y), time &
\{"POINT":[x,y],"duration": time*1000\} \\

\texttt{swipe} & coordinate = (x1, y1), coordinate2 = (x2, y2) &
\{"POINT":[x1,y1],"to":\newline direction\}\newline
\myit{Note: direction is computed from two points} \\

\texttt{type} & text &
\{"TYPE":text\} \\

\texttt{system\_button} & button = Back / Home / Enter &
\{"PRESS":BACK/HOME/ENTER\} \\

\texttt{terminate} & None &
\{"STATUS":"finish"\} \\

\texttt{wait} & time &
\{"duration":time*1000\} \\
\bottomrule
\end{tabularx}
\end{table}

\subsubsection{Hyperparameters}

We adopt the same hyperparameter settings as used in Qwen2.5-VL-7B for fair comparison, as summarized in Table~\ref{tab:qwen_hyperparameters}.
\begin{table}[h]
\centering
\caption{Inference hyperparameters for Qwen2.5-VL-7B.}
\begin{tabularx}{\textwidth}{c c X}
\toprule
\textbf{Parameter} & \textbf{Default Value} & \textbf{Description} \\
\midrule
\texttt{do\_sample}         & True   & Whether to use sampling (replaces greedy) \\
\texttt{top\_p}             & 0.01   & Nucleus sampling threshold \\
\texttt{top\_k}             & 1      & Top-k sampling limit \\
\texttt{temperature}        & 0.01   & Controls sampling randomness \\
\texttt{repetition\_penalty}& 1.0    & Penalty factor for repetition \\
\texttt{max\_new\_tokens}   & 2048   & Maximum number of new tokens to generate \\
\bottomrule
\end{tabularx}
\label{tab:qwen_hyperparameters}
\end{table}

\subsection{UI-TARS}

\subsubsection{Data example}

\begin{tcolorbox}[breakable, colback=black!5!white,colframe=black!75!black,title=UI-TARS Data Example]
\tcbsubtitle{System Message}
You are a helpful assistant.
\tcbsubtitle{User}
You are a GUI agent. You are given a task and your action history, with screenshots. You need to perform the next action to complete the task.\\

\textbf{Output Format}\\

\texttt{Thought: \ldots}\\
\texttt{Action: \ldots}\\

\vspace{1em}
\textbf{Action Space}\\
\texttt{click(start\_box='\textless|box\_start|\textgreater(x1,y1)\textless|box\_end|\textgreater')}\\
\texttt{long\_press(start\_box='\textless|box\_start|\textgreater(x1,y1)\textless|box\_end|\textgreater', time='')}\\
\texttt{type(content='')}\\
\texttt{scroll(direction='down or up or right or left')}\\
\texttt{press\_back()}\\
\texttt{press\_home()}\\
\texttt{wait()}\\
\texttt{finished()} \# Submit the task regardless of whether it succeeds or fails.\\

\vspace{1em}
\textbf{Note}\\
- Use English in Thought part.\\
- Summarize your next action (with its target element) in one sentence in Thought part.\\

\textbf{User Instruction}\\
\texttt{[user\_request]}
\tcbsubtitle{User}
[history\_screenshot]
\tcbsubtitle{Assistant}
[history\_thought\_and\_action]
\tcbsubtitle{User}
[current\_screenshot]
\tcbsubtitle{Assistant(included only when low\_lew\_instruction is defined)}
Thought: [low\_lew\_instruction]

Action:
\tcbsubtitle{Assistant}
[thought\_and\_action]
\label{eg_uitars}
\end{tcolorbox}

\subsubsection{Action Space Mapping}

Table~\ref{tab:uitars_mapping} shows the action space mapping from UI-TARS to the standardized representation. Since UI-TARS and AgentCPM-GUI define scroll directions oppositely, the direction must be reversed during conversion.

\begin{table}[htbp]
\centering
\caption{Action space mapping from UI-TARS to AgentCPM-GUI.}
\label{tab:uitars_mapping}
\begin{tabularx}{\textwidth}{l X X}
\toprule
\textbf{UI-TARS} & \textbf{Input Format} & {\normalfont\textbf{AgentCPM-GUI}} \\
\midrule
\texttt{click(...)} & start\_box with (x, y) & \texttt{\{"POINT":[x,y]\}} \\

\texttt{long\_press(...)} & start\_box with (x, y), time=`ms' (optional) &
\texttt{\{"POINT":[x,y],"duration": time (default 1000)\}} \\

\texttt{type(...)} & content=`text' & \texttt{\{"TYPE":text\}} \\

\texttt{scroll(...)} & direction=`up/down/left/right' &
\parbox[t]{\linewidth}{%
  \hspace*{0pt}\texttt{\{"POINT":[500,500],} \\
  \hspace*{0pt}\texttt{"to":reversed direction\}}
} \newline
\textit{Note: direction is reversed (e.g., up $\rightarrow$ down)} \\

\texttt{press\_back()} & - & \texttt{\{"PRESS":BACK\}} \\

\texttt{press\_home()} & - & \texttt{\{"PRESS":HOME\}} \\

\texttt{wait()} & - & \texttt{\{"duration":200\}} \\

\texttt{finished()} & - & \texttt{\{"STATUS":"finish"\}} \\
\bottomrule
\end{tabularx}
\end{table}

\subsection{OS-ATLAS}

\subsubsection{Data example}

\begin{tcolorbox}[breakable, colback=black!5!white,colframe=black!75!black,title=OS-ATLAS Data Example]
\tcbsubtitle{System Message}
You are a helpful assistant.
\tcbsubtitle{User}
{\normalsize	
You are a foundational action model capable of automating tasks across various digital environments, including desktop systems like Windows, macOS, and Linux, as well as mobile platforms such as Android and iOS. You also excel in web browser environments. You will interact with digital devices in a human-like manner: by reading screenshots, analyzing them, and taking appropriate actions.\\

Your expertise covers two types of digital tasks:\\
\ \ \ \ - Grounding: Given a screenshot and a description, you assist users in locating elements mentioned. Sometimes, you must infer which elements best fit the description when they aren't explicitly stated.\\
\ \ \ \ - Executable Language Grounding: With a screenshot and task instruction, your goal is to determine the executable actions needed to complete the task.\\

You are now operating in Executable Language Grounding mode. Your goal is to help users accomplish tasks by suggesting executable actions that best fit their needs. Your skill set includes both basic and custom actions:\\

1. Basic Actions\\
Basic actions are standardized and available across all platforms. They provide essential functionality and are defined with a specific format, ensuring consistency and reliability.\\
Basic Action 1: CLICK\\
\ \ \ \ - purpose: Click at the specified position.\\
\ \ \ \ - format: \texttt{CLICK \textless point\textgreater[[x-axis, y-axis]]\textless/point\textgreater}\\
\ \ \ \ - example usage: \texttt{CLICK \textless point\textgreater[[101, 872]]\textless/point\textgreater}\\

Basic Action 2: TYPE\\
\ \ \ \ - purpose: Enter specified text at the designated location.\\
\ \ \ \ - format: \texttt{TYPE [input text]}\\
\ \ \ \ - example usage: \texttt{TYPE [Shanghai shopping mall]}\\

Basic Action 3: SCROLL\\
\ \ \ \ - purpose: Scroll in the specified direction.\\
\ \ \ \ - format: \texttt{SCROLL [direction (UP/DOWN/LEFT/RIGHT)]}\\
\ \ \ \ - example usage: \texttt{SCROLL [UP]}\\

2. Custom Actions\\
Custom actions are unique to each user's platform and environment. They allow for flexibility and adaptability, enabling the model to support new and unseen actions defined by users. These actions extend the functionality of the basic set, making the model more versatile and capable of handling specific tasks.\\

Custom Action 1: LONG\_PRESS\\
\ \ \ \ - purpose: Long press at the specified position.\\
\ \ \ \ - format: \texttt{LONG\_PRESS \textless point\textgreater[[x-axis, y-axis]]\textless/point\textgreater}\\
\ \ \ \ - example usage: \texttt{LONG\_PRESS \textless point\textgreater[[101, 872]]\textless/point\textgreater}\\

Custom Action 2: PRESS\_BACK\\
\ \ \ \ - purpose: Press a back button to navigate to the previous screen.\\
\ \ \ \ - format: \texttt{PRESS\_BACK}\\
\ \ \ \ - example usage: \texttt{PRESS\_BACK}\\

Custom Action 3: PRESS\_HOME\\
\ \ \ \ - purpose: Press a home button to navigate to the home page.\\
\ \ \ \ - format: \texttt{PRESS\_HOME}\\
\ \ \ \ - example usage: \texttt{PRESS\_HOME}\\

Custom Action 4: PRESS\_RECENT\\
\ \ \ \ - purpose: Press the recent button to view or switch between recently used applications.\\
\ \ \ \ - format: \texttt{PRESS\_RECENT}\\
\ \ \ \ - example usage: \texttt{PRESS\_RECENT}\\

Custom Action 5: WAIT\\
\ \ \ \ - purpose: Wait for the screen to load.\\
\ \ \ \ - format: \texttt{WAIT}\\
\ \ \ \ - example usage: \texttt{WAIT}\\

Custom Action 6: COMPLETE\\
\ \ \ \ - purpose: Indicate the task is finished.\\
\ \ \ \ - format: \texttt{COMPLETE}\\
\ \ \ \ - example usage: \texttt{COMPLETE}\\

In most cases, task instructions are high-level and abstract. Carefully read the instruction and action history, then perform reasoning to determine the most appropriate next action. Ensure you strictly generate two sections: Thoughts and Actions.\\
\textbf{Thoughts}: Clearly outline your reasoning process for current step.\\
\textbf{Actions}: Specify the actual actions you will take based on your reasoning.\\

Your current task instruction, action history, and associated screenshot are as follows:\\
Screenshot:[current\_screenshot]

Task: [user\_request] You need to: [low\_lew\_instruction](included only when low\_lew\_instruction is defined)

History: 

[history\_low\_lew\_instruction](included only when low\_lew\_instruction is defined)
}
\tcbsubtitle{Assistant}
[thought\_and\_action]
\label{eg_osatlas}
\end{tcolorbox}

\subsubsection{Action Space Mapping}

Table~\ref{tab:osatlas_mapping} shows the action space mapping from OS-ATLAS to the standardized representation. When evaluating the AndroidControl-Low setting, we found that the model's predicted scroll direction is often opposite to that indicated in the low-level instruction. Therefore, the scroll direction is reversed during evaluation.

\begin{table}[htbp]
\centering
\caption{Action space mapping from OS-Atlas to AgentCPM-GUI.}
\label{tab:osatlas_mapping}
\begin{tabularx}{\textwidth}{l l Y}
\toprule
\textbf{OS-Atlas} & \textbf{Input Format} & {\normalfont\textbf{AgentCPM-GUI}} \\
\midrule
\texttt{CLICK} & [[x, y]] & \{"POINT":[x,y]\} \\

\texttt{LONG\_PRESS} & [[x, y]] & \{"POINT":[x,y],"duration":1000\} \\

\texttt{TYPE} & [text] & \{"TYPE":text\} \\

\texttt{SCROLL} & [direction] & \{"POINT":[500,500],"to":direction\} \newline
\myit{Note: if use\_low\_instruction is True, direction is reversed:\;up$\leftrightarrow$down,\;left$\leftrightarrow$right}
\\
\texttt{PRESS\_BACK} & - & \{"PRESS":BACK\} \\

\texttt{PRESS\_HOME} & - & \{"PRESS":HOME\} \\

\texttt{PRESS\_RECENT} & - & \{"PRESS":RECENT\} \\

\texttt{WAIT} & - & \{"duration":200\} \\

\texttt{COMPLETE} & - & \{"STATUS":"finish"\} \\
\bottomrule
\end{tabularx}
\end{table}

\subsection{OS-Genesis}

\subsubsection{Data Example}

For the GUI-Odyssey, AITZ, and CAGUI benchmarks, we construct evaluation prompts following the format described in \nameref{eg_osgensis}. For AndroidControl, we adopt the official evaluation code provided in the benchmark's GitHub repository.

\begin{tcolorbox}[breakable, colback=black!5!white,colframe=black!75!black,title=OS-Genesis Data Example]
\tcbsubtitle{System Message}
You are a helpful assistant.
\tcbsubtitle{User}
You are a GUI task expert, I will provide you with a high-level instruction, an action history, a screenshot with its corresponding accessibility tree.\\

High-level instruction: [user\_request]

Action history: 

Accessibility tree: 

Please generate the low-level thought and action for the next step.

\tcbsubtitle{Assistant}
[thought\_and\_action]
\label{eg_osgensis}
\end{tcolorbox}

\subsubsection{Action Space Mapping}

Table~\ref{tab:osgensis_mapping} shows the action space mapping from OS-Genesis to the standardized representation. Similar to OS-ATLAS, the predicted scroll direction on AndroidControl-Low is often opposite to the instruction, and is therefore reversed during evaluation.

\begin{table}[htbp]
\centering
\caption{Action space mapping from OS-Genesis to AgentCPM-GUI.}
\label{tab:osgensis_mapping}
\begin{tabularx}{\textwidth}{l l Y}
\toprule
\textbf{OS-Genesis} & \textbf{Input Fields} & {\normalfont\textbf{AgentCPM-GUI}} \\
\midrule
\texttt{type} & text & \{"TYPE":text\} \\

\texttt{click} & x, y & \{"POINT":[x,y]\} \\

\texttt{long\_press} & x, y & \{"POINT":[x,y],"duration":1000\} \\

\texttt{dismiss} & x, y & \{"POINT":[x,y]\} \\

\texttt{get\_text} & x, y & \{"POINT":[x,y]\} \\

\texttt{navigate\_home} & - & \{"PRESS":HOME\} \\

\texttt{navigate\_back} & - & \{"PRESS":BACK\} \\

\texttt{scroll} & direction & \{"POINT":[500,500],"to":direction\} \newline
\myit{Note: If use\_low\_instruction is True, direction is reversed:\;up$\leftrightarrow$down,\;left$\leftrightarrow$right}
\\
\texttt{wait} & - & \{"duration":200\} \\
\bottomrule
\end{tabularx}
\end{table}

\subsection{OdysseyAgent}

\subsubsection{Data example}

Following the official implementation, OdysseyAgent's input consists of the current instruction along with a history of images and their associated actions. 

\begin{tcolorbox}[colback=black!5!white,colframe=black!75!black,title=OdysseyAgent Data Example]
\tcbsubtitle{System Message}
You are a helpful assistant.
\tcbsubtitle{User}
Picture 1: <img>{image\_path}</img>\\
I'm looking for guidance on how to [instruction]\\
Previous screenshots: <img>image-history: {image\_path}</img>\\
Previous Actions: 1. [Action 1]\\
2. [Action 2].\\
...
\tcbsubtitle{Assistant}
[Action]
\label{eg_odyssey}
\end{tcolorbox}

\subsubsection{Action Space Mapping}

Table~\ref{tab:odyssey_mapping} shows the action space mapping from OdysseyAgent to the standardized representation. The output format of OdysseyAgent is largely compatible with AgentCPM-GUI. The only exception is the \texttt{RECENT} action, which is not part of the AgentCPM-GUI action space and is therefore ignored during evaluation.

\begin{table}[htbp]
\centering
\caption{Action space mapping from OdysseyAgent to AgentCPM-GUI.}
\begin{tabularx}{\textwidth}{l l >{\ttfamily\raggedright\arraybackslash}X}
\toprule
\textbf{OdysseyAgent} & \textbf{Input Fields} & \normalfont\textbf{AgentCPM-GUI} \\
\midrule
\texttt{CLICK}                & x, y                          & \{"POINT":{[}x,y{]}\} \\
\texttt{LONG\_PRESS}          & x, y                    & \{"POINT":{[}x,y{]},"duration":1000\} \\
\texttt{SCROLL}                &  direction               & \{"POINT":{[}500,500{]},"to":direction\} \\
\texttt{TYPE}                 & text                     & \{"TYPE":text\} \\
\texttt{HOME}       & -                     & \{"PRESS":HOME\} \\
\texttt{BACK}       & -                     & \{"PRESS":BACK\} \\
\texttt{COMPLETE}            & -                             & \{"STATUS":"finish"\} \\
\texttt{IMPOSSIBLE}           & -                           & \{"STATUS":"impossible"\} \\
\bottomrule
\end{tabularx}

\label{tab:odyssey_mapping}
\end{table}

\subsubsection{Hyperparameters}

We follow the original implementation for inference, enabling the \texttt{image\_history} option to incorporate temporal context. Specifically, we store the last $4$ actions and their corresponding images. The inference is conducted with the torch seed set to $1234$ and the random seed set to $2020$ to ensure reproducibility.

\subsection{Aguvis-7B}

\subsubsection{Data Example}

\begin{tcolorbox}[colback=black!5!white,colframe=black!75!black,title=Aguvis Data Example]
\tcbsubtitle{System Message}
You are a GUI agent. You are given a task and a screenshot of the screen. You need to perform a series of pyautogui actions to complete the task.

You have access to the following functions:\\
- \{"name": "mobile.swipe", "description": "Swipe on the screen", "parameters": \{"type": "object", "properties": \{"from\_coord": \{"type": "array", "items": \{"type": "number"\}, "description": "The starting coordinates of the swipe"\}, "to\_coord": \{"type": "array", "items": \{"type": "number"\}, "description": "The ending coordinates of the swipe"\}\}, "required": ["from\_coord", "to\_coord"]\}\}\\
- \{"name": "mobile.home", "description": "Press the home button"\}\\
- \{"name": "mobile.back", "description": "Press the back button"\}\\
- \{"name": "mobile.wait", "description": "wait for the change to happen", "parameters": \{"type": "object", "properties": \{"seconds": \{"type": "number", "description": "The seconds to wait"\}\}, "required": ["seconds"]\}\}\\
- \{"name": "mobile.long\_press", "description": "Long press on the screen", "parameters": \{"type": "object", "properties": \{"x": \{"type": "number", "description": "The x coordinate of the long press"\}, "y": \{"type": "number", "description": "The y coordinate of the long press"\}\}, "required": ["x", "y"]\}\}\\
- \{"name": "mobile.open\_app", "description": "Open an app on the device", "parameters": \{"type": "object", "properties": \{"app\_name": \{"type": "string", "description": "The name of the app to open"\}\}, "required": ["app\_name"]\}\}
\tcbsubtitle{User}
Please generate the next move according to the ui screenshot, instruction and previous actions.

Instruction: [Instruction]

Previous actions:
[previous\_actions]
\tcbsubtitle{Assistant}
[thought and Action]
\label{eg_aguvis}
\end{tcolorbox}

\begin{table}[htbp]
\centering
\caption{Action space mapping from Aguvis to AgentCPM-GUI.}
\begin{tabularx}{\textwidth}{l l >{\ttfamily\raggedright\arraybackslash}X}
\toprule
\textbf{Aguvis} & \textbf{Input Fields} & \normalfont\textbf{AgentCPM-GUI} \\
\midrule
\texttt{pyautogui.click}                       & x, y         &\{"POINT":{[}x*1000,y*1000{]}\} \\
\texttt{mobile.long\_press}                & x, y       & \{"POINT":{[}x*1000,y*1000{]},\newline"duration":1000\} \\
\texttt{pyautogui.scroll()/hscroll()}                       & direction  & \{"POINT":{[}500,500{]},"to": direction\}\newline\myit{Note: \texttt{scroll} performs vertical, and \texttt{hscroll} performs horizontal swipes} \\
\texttt{pyautogui.write}                        & text         & \texttt{\{"TYPE":text\}} \\
\texttt{mobile.home()/}              & -          & \{"PRESS":HOME\} \\
\texttt{mobile.back()}              & -          & \{"PRESS":BACK\} \\
\texttt{mobile.terminate()}                   & -            & \{"STATUS":"finish"\} \\
\texttt{mobile.open\_app}                   & app\_name   & - \\
\texttt{mobile.wait}                        & {[}time{]}             & \{"duration":3000\} \\
\bottomrule
\end{tabularx}
\label{tab:aguvis_mapping}
\end{table}

\subsubsection{Action Space Mapping}

Table~\ref{tab:aguvis_mapping} shows the action space mapping from Aguvis to the standardized representation. All coordinates in Aguvis are in the range $[0,1]$ and are scaled accordingly during conversion. Swipe actions are mapped following the definition in the \texttt{pyautogui} package. Since AgentCPM-GUI does not include an "open app" action, it is ignored during evaluation.

\subsubsection{Hyperparameters}
The hyper parameters are the same as the origin implementation. To be specific, we choose "self-plan" mode during inference, with temperature set as $0$ and generate only $1024$ new max tokens. Historical actions are not included during inference, as their inclusion leads to abnormal model behavior.

\section{CAGUI Benchmark}

\subsection{CAGUI\_Grounding}

We provide examples from the three tasks that constitute the grounding benchmark, each containing 1,500 samples. The Text2Bbox and Bbox2Text tasks are based on the same dataset. Each bounding box is defined by four absolute coordinates in the format <$x_{\min}$, $y_{\min}$, $x_{\max}$, $y_{\max}$>, with the origin located at the top-left corner of the screen.

\label{appendix:CAGUI_Grounding}


\label{appendix:CAGUI_Grounding_Task_Examples}
\begin{tcolorbox}[colback=black!5!white,colframe=black!75!black,title=Text2Point Data Examples]
\tcbsubtitle{Text}
\begin{CJK*}{UTF8}{gbsn}
QQ音乐
\end{CJK*}
\tcbsubtitle{Bounding Box}
<643, 462, 849, 744>
\tcbsubtitle{Prompt of AgentCPM-GUI}
\begin{CJK*}{UTF8}{gbsn}
你是一个GUI组件定位的专家，擅长输出图片上文本对应的坐标。你的任务是根据给定的GUI截图和图中某个文本输出该文本的坐标。   输入：屏幕截图，文本描述   输出：文本的相对坐标的中心点,{POINT:[...,...]}为格式
\end{CJK*}
\end{tcolorbox}

\begin{tcolorbox}[colback=black!5!white,colframe=black!75!black,title=Bbox2Text Data Examples]
\tcbsubtitle{Bounding Box}
<60, 120, 132, 192>
\tcbsubtitle{Bounding Box}
\begin{CJK*}{UTF8}{gbsn}
返回
\end{CJK*}
\tcbsubtitle{Prompt of AgentCPM-GUI}
\begin{CJK*}{UTF8}{gbsn}
你是一个GUI组件文字识别的专家，擅长根据组件的边界框（bounding box）描述输出对应的文字。你的任务是根据给定的GUI截图和图中某个组件的边界框输出组件的中的文字。   输入：屏幕截图，边界框的坐标   输出：组件中的文本
\end{CJK*}
\end{tcolorbox}

\begin{tcolorbox}[colback=black!5!white,colframe=black!75!black,title=Fun2Point Data Examples]
\tcbsubtitle{Function}
\begin{CJK*}{UTF8}{gbsn}UI元素是一个菜单按钮。其主要功能是弹出一个菜单面板，允许用户选择不同的功能选项。通常可以通过点击该按钮触发，点击后会展示一个下拉或侧滑菜单，用户可以在其中进行进一步操作，例如切换功能页面或设置选项。
\end{CJK*}
\tcbsubtitle{Bounding Box}
<1061, 2424, 1159, 2522>
\tcbsubtitle{Prompt of AgentCPM-GUI}
\begin{CJK*}{UTF8}{gbsn}你是一个GUI组件定位的专家，擅长根据组件的功能描述输出对应的坐标。你的下一步操作是根据给定的GUI截图和图中某个组件的功能描述点击组件的中心位置。坐标为相对于屏幕左上角位原点的相对位置，并且按照宽高比例缩放到0～1000    输入：屏幕截图，功能描述    输出：点击操作，以{POINT:[...,...]}为格式，其中不能存在任何非坐标字符
\end{CJK*}
\end{tcolorbox}

\subsection{CAGUI\_Agent}


We present examples of our dataset tasks, each consisting of a query, a screenshot, and the corresponding answer operation. The system prompt used to evaluate AgentCPM-GUI is also included. In total, the benchmark comprises 600 tasks, which together contain 4,516 single-step images. During evaluation, inputs to AgentCPM-GUI follow the standard chat format. Each user message contains both the task query and the associated screenshot, structured as a list with two elements: a text string formatted as "{\begin{CJK*}{UTF8}{gbsn}{<Question>\{query\}</Question>\textbackslash n当前屏幕截图：}\end{CJK*}}" and the corresponding image.

\begin{tcolorbox}[breakable, colback=black!5!white,colframe=black!75!black,title=Agent Data Examples]
\tcbsubtitle{Query}
\begin{CJK*}{UTF8}{gbsn}请优酷视频根据我的历史记录播放7天内观看超过60\%的短视频。
\end{CJK*}
\tcbsubtitle{Operation}
Action Type: Click

Action Detail: [0.13, 0.61]
\tcbsubtitle{System Prompt of AgentCPM-GUI}
\begin{CJK*}{UTF8}{gbsn}

\noindent\textbf{\# Role} \\
你是一名熟悉安卓系统触屏GUI操作的智能体，将根据用户的问题，分析当前界面的GUI元素和布局，生成相应的操作。 \\

\noindent\textbf{\# Task} \\
针对用户问题，根据输入的当前屏幕截图，输出下一步的操作。 \\

\noindent\textbf{\# Rule} \\
- 以紧凑JSON格式输出 \\
- 输出操作必须遵循Schema约束 \\

\noindent\textbf{\# Schema}
\begin{verbatim}
{
  "type": "object",
  "description": "执行操作并决定当前任务状态",
  "additionalProperties": false,
  "properties": {
    "thought": {
      "type": "string",
      "description": "智能体的思维过程"
    },
    "POINT": {
      "$ref": "#/$defs/Location",
      "description": "点击屏幕上的指定位置"
    },
    "to": {
      "description": "移动，组合手势参数",
      "oneOf": [
        {
          "enum": [
            "up",
            "down",
            "left",
            "right"
          ],
          "description": "从当前点（POINT）出发，执行滑动手势操作，方向包括向上、向下、向左、向右"
        },
        {
          "$ref": "#/$defs/Location",
          "description": "移动到某个位置"
        }
      ]
    },
    "duration": {
      "type": "integer",
      "description": "动作执行的时间或等待时间，毫秒",
      "minimum": 0,
      "default": 200
    },
    "PRESS": {
      "type": "string",
      "description": "触发特殊按键，HOME为回到主页按钮，BACK为返回按钮，ENTER为回车按钮",
      "enum": [
        "HOME",
        "BACK",
        "ENTER"
      ]
    },
    "TYPE": {
      "type": "string",
      "description": "输入文本"
    },
    "STATUS": {
      "type": "string",
      "description": "当前任务的状态。特殊情况：satisfied，无需操作；impossible，任务无法完成；interrupt，任务中断；need_feedback，需要用户反馈；",
      "enum": [
        "continue",
        "finish",
        "satisfied",
        "impossible",
        "interrupt",
        "need_feedback"
      ],
      "default": "continue"
    }
  },
  "$defs": {
    "Location": {
      "type": "array",
      "description": "坐标为相对于屏幕左上角位原点的相对位置，并且按照宽高比例缩放到0～1000，数组第一个元素为横坐标x，第二个元素为纵坐标y",
      "items": {
        "type": "integer",
        "minimum": 0,
        "maximum": 1000
      },
      "minItems": 2,
      "maxItems": 2
    }
  }
}
\end{verbatim}

\end{CJK*}
\end{tcolorbox}

\newpage

\section{Case Study}

We demonstrate GUI agent tasks on a real Xiaomi 12S device running MIUI 14.0.11. All interactions with the graphical interface are carried out via ADB control based on the AgentCPM-GUI's predicted actions. The original input and output were in Chinese and translated into English.

\begin{figure}[htpb]
    \centering
  \includegraphics[width=0.9\linewidth]{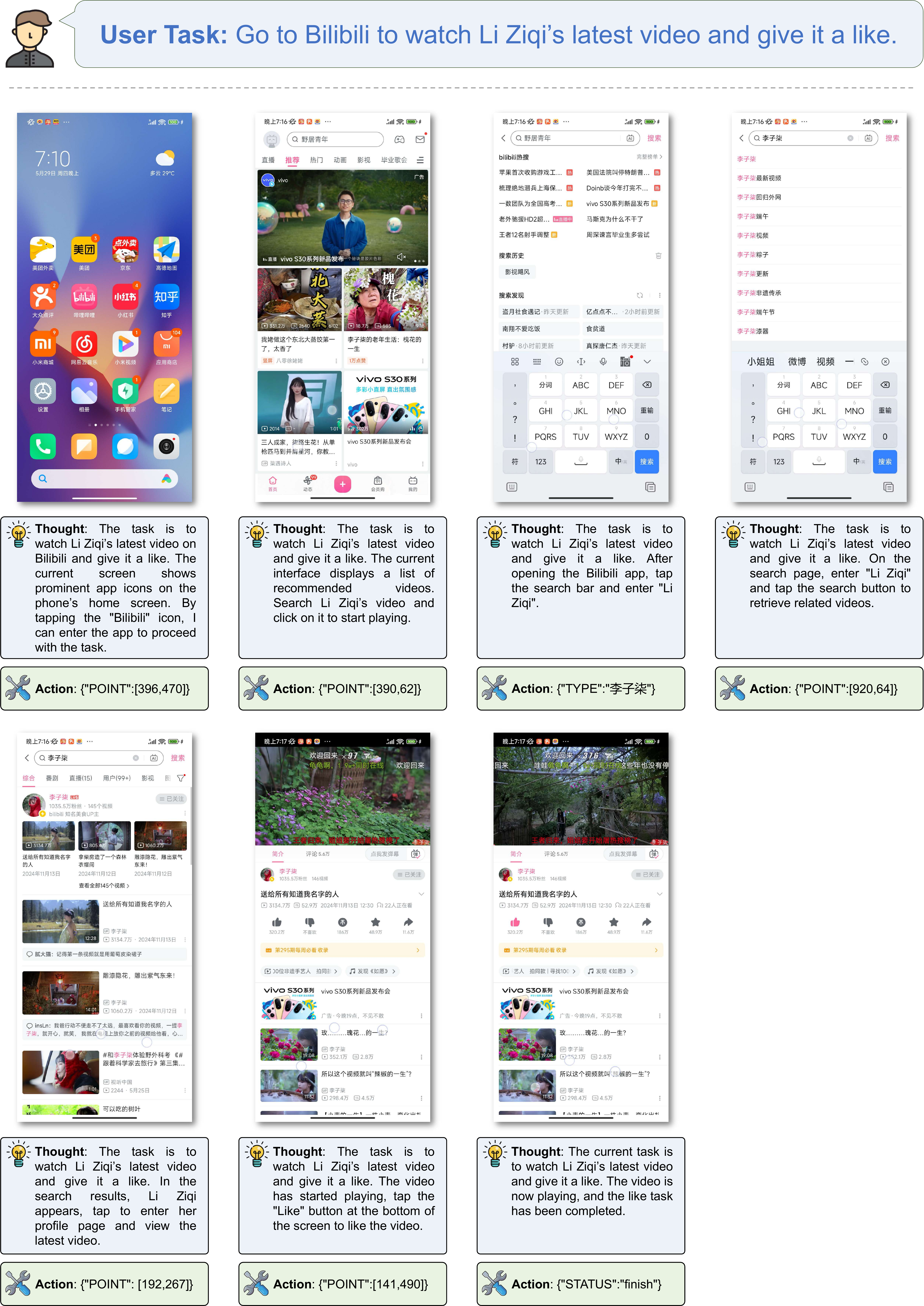}
    \caption{A demo case on the BiliBili.}
    \label{fig:case_study_liziqi}
\end{figure}

\begin{figure}[htpb]
    \centering
  \includegraphics[width=0.9\linewidth]{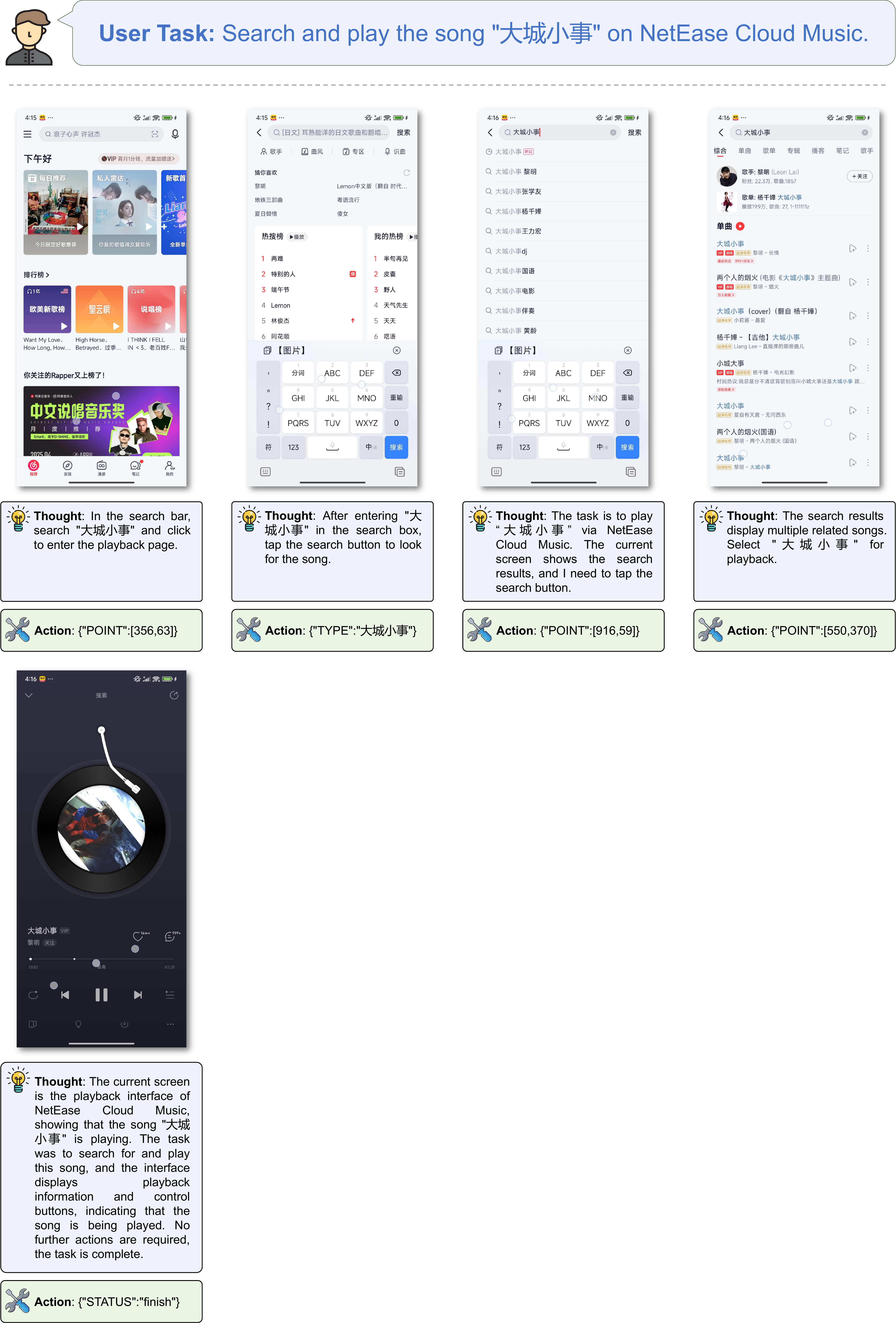}
    \caption{A demo case on the NetEase Cloud Music.}
    \label{fig:case_study_NetEase}
\end{figure}

\end{document}